\documentclass[preprint]{elsarticle}

\usepackage{hyperref}
\usepackage{setspace}

\usepackage{geometry}
\usepackage{amsmath}
\usepackage{multirow}
\usepackage{amssymb}
\usepackage{wasysym}

\usepackage{setspace}
\usepackage{changes}
\usepackage{makecell}
\usepackage{threeparttable} 
\usepackage{booktabs} 

\journal{} 

\biboptions{authoryear}

\makeatletter
\renewcommand{\ps@pprintTitle}{%
  \let\@oddhead\@empty
  \let\@evenhead\@empty
  \def\@oddfoot{}%
  \let\@evenfoot\@oddfoot}
\makeatother

\begin{document}

\begin{frontmatter}

\title{From Noisy Historical Maps to Time-Series Oil Palm Mapping Without Annotation in Malaysia and Indonesia (2020–2024)}


\author[SYSU]{Nuttaset Kuapanich}
\author[SYSU,NSCCSZ]{Juepeng Zheng\fnref{myfootnote}}
\author[NEU]{Bohan Shi}
\author[SYSU]{Jiaying Liu}
\author[SYSU]{Jiayin Jiang}
\author[SYSU]{Jiatao Huang} 
\author[SYSU]{Shenghan Tan}
\author[THU]{Qingmei Li}
\author[NSCCSZ,THU]{Haohuan Fu}

\address[SYSU]{School of Artificial Intelligence, Sun Yat-Sen University, Zhuhai, China.}
\address[NSCCSZ]{National Supercomputing Center in Shenzhen, Shenzhen, China.}
\address[NEU]{Khoury College of Computer Sciences, Northeastern University, Boston, United States of America.}
\address[THU]{Tsinghua Shenzhen International Graduate School, Tsinghua University, Shenzhen, China.}
\fntext[myfootnote]{Corresponding author: zhengjp8@mail.sysu.edu.cn}


\begin{abstract}
Accurate monitoring of oil palm plantations is critical for balancing economic development with environmental conservation in Southeast Asia. However, existing plantation maps often suffer from low spatial resolution and a lack of recent temporal coverage, impeding effective surveillance of rapid land-use changes. In this study, we propose a deep learning framework to generate 10-meter resolution oil palm plantation maps for Indonesia and Malaysia from 2020 to 2024, utilizing Sentinel-2 imagery without requiring new manual annotations. To address the resolution mismatch between coarse 100-meter historical labels and 10-meter imagery, we employ a U-Net architecture optimized with Determinant-based Mutual Information (DMI). This approach effectively mitigates the influence of label noise. We validated our method against 2,058 manually verified points, achieving overall accuracies of 70.64\%, 63.53\%, and 60.06\% for the years 2020, 2022, and 2024, respectively. Our comprehensive analysis reveals that oil palm coverage in the region peaked in 2022 before experiencing a decline in 2024. Furthermore, land cover transition analysis highlights a concerning trajectory of plantation expansion into flooded vegetation areas, despite a general stabilization in rotations with other crop types. These high-resolution maps provide essential data for monitoring sustainability commitments and deforestation dynamics in the region, and the generated datasets are made publicly available at \url{https://doi.org/10.5281/zenodo.17768444}.
\end{abstract}


\end{frontmatter}


\section{Introduction}

Global demand for palm oil has risen sharply from 2 million tons in 1970 to nearly 80 million tons in 2020, which together produce over 84\% of global output and hold two-thirds of the world’s oil palm cultivation area in Indonesia and Malaysia, to rapidly expand plantation land to meet market needs \citep{owid-palm-oil}. While oil palm provides farmers with higher income and supports rural economic development, this expansion has generated substantial environmental pressure. Both countries have experienced extensive deforestation, with RSPO-certified concessions in Indonesia overlapping habitats of critically endangered species and retaining only a small fraction of original forest cover \citep{defor_in}, and Malaysia losing nearly 6 million hectares of forest between 2010 and 2018 \citep{impact_ma}. Large-scale plantation development also accelerates carbon emissions, particularly in peat-rich regions of Indonesia where land conversion releases large amounts of stored carbon \citep{ghg_ma, ghg_in}. These intertwined economic and ecological dynamics highlight the urgent need for accurate, temporally consistent mapping of oil palm plantations to support sustainable land management and monitoring.

In recent decades, monitoring oil palm plantations has traditionally relied on extensive field-based surveys, requiring substantial labor, financial investment, and logistical effort across large and often inaccessible tropical regions. To reduce this burden, the remote sensing community increasingly adopted machine learning techniques to map oil palm distribution using multispectral satellite imagery from platforms such as Sentinel-2 and Landsat \citep{ml_thai, ml_in}. Early studies demonstrated the value of spectral–textural features combined with classifiers such as Random Forest, XGBoost, and Classification and Regression Trees (CART) \citep{ml_ieee, ml_in}. These models proved effective in distinguishing oil palm from other woody vegetation, shrublands, and agricultural crops, and were widely used for generating regional plantation maps and conducting long-term time-series analyses of plantation expansion \citep{ml_annu}. The adoption of these traditional machine learning approaches markedly reduced the dependence on costly field surveys and enabled large-area assessments that were previously infeasible.

As neural networks advanced, deep learning methods rapidly became the dominant paradigm for oil palm mapping due to their ability to automatically learn hierarchical spectral–spatial representations. Convolutional neural networks (CNNs) and semantic segmentation architectures (such as U-Net, DeepLab, and fully convolutional networks) have consistently outperformed traditional machine learning models in delineating plantation boundaries, identifying palm crowns, and capturing fine-scale plantation structures \citep{dl_ml}. For instance, Adrià et al. demonstrated that DeepLabv3+ achieves significantly higher accuracy than Random Forest classifiers when detecting oil palm plantations using high-resolution imagery. More recently, object detection and instance segmentation frameworks have been applied to count individual palms, estimate stand density, and monitor plantation age structure. However, the performance of deep learning models critically depends on access to large, high-quality annotated datasets, which poses a major bottleneck for scaling these methods to national or continental extents. Rodríguez et al. \citep{dl_al} manually annotated approximately 3.4 × 10$^6$ oil palm instances, while Putra et al. \citep{dl_obj} created nearly 100,000 manually labeled bounding boxes across 507 images—both requiring substantial human effort, domain expertise, and computational resources. This annotation burden significantly limits the development of more advanced deep learning models for oil palm mapping, especially in regions where ground truth data are scarce or inconsistently available.



In remote sensing applications, numerous approaches have been proposed to reduce the reliance on manual annotation by leveraging existing map products as supervisory signals. Because these ancillary maps are often outdated, spatially coarse, or thematically inconsistent, recent studies have increasingly adopted noisy-label learning strategies to mitigate label uncertainty. Techniques such as pseudo-label refinement, consistency-based learning, and k-nearest-neighbor–based noise correction have shown promising performance in large-scale land cover classification tasks, particularly when training with imperfect labels \citep{lc_1, lc_2}. More advanced frameworks further incorporate label smoothing, uncertainty estimation, or teacher–student models to enhance robustness against mislabeled pixels. However, most of these methods are tailored to broad land cover or land use mapping \citep{lc_3}, where class semantics are relatively generic and spatial patterns are less crop-specific. In contrast, oil palm plantations exhibit distinct phenological cycles and spatial structures that are not explicitly represented in existing global or regional maps, leading to substantial label noise when these products are used as supervision. Consequently, despite substantial progress in remote sensing segmentation and noisy-label learning, accurate large-scale oil palm mapping remains challenging, and current products still suffer from limited classification accuracy. This gap highlights the urgent need for dedicated, noise-aware approaches specifically designed for oil palm plantation detection in complex tropical landscapes.


\begin{figure}[t]
    \centering
    \includegraphics[width=\linewidth]{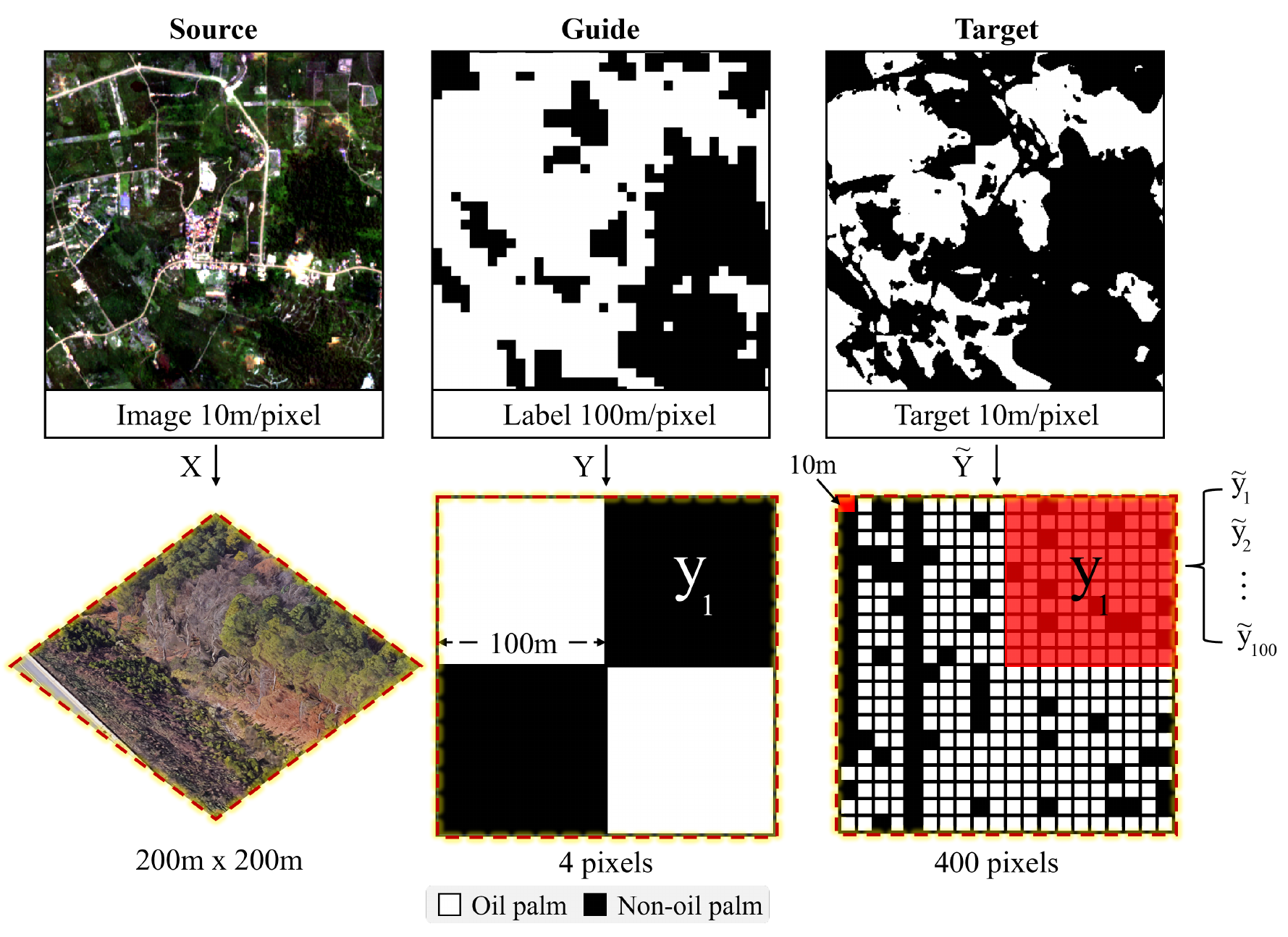}
    \caption{Precision comparison of oil palm plantation maps between 100 meter and 10 meter resolution}
    \label{fig:intro}
\end{figure}

To the best of our knowledge, available oil palm plantation maps predominantly have spatial resolutions of 30 and 100 meters and cover the year range from 2001 to 2019, with no available 10 meter resolution maps from 2020 onwards at the country scale (including Indonesia and Malaysia) \citep{ml_annu, ml_thai, dl_al}. This temporal and spatial resolution gap is particularly significant, as higher resolution mapping is crucial for accurate plantation boundary delineation and monitoring, as illustrated in Fig.  \ref{fig:intro}.  Therefore, this study aims to develop comprehensive 10 meter resolution oil palm plantation mapping products for Malaysia and Indonesia for the period 2020-2024 without requiring additional manual labeling.

Therefore, the contributions of this paper mainly include the following three points:

\begin{enumerate}
    \item We propose a new noise learning method for oil palm plantation mapping that can be completed without any manual labels. Specifically, we employ a U-Net architecture optimized with Determinant-based Mutual Information (DMI).
   
    \item We develop a set of maps for 2020-2024 and manually confirmed 2,058 validation points, achieving an Overall Accuracy (OA) of 70.64\% in 2020, 63.53\% in 2022 and 60.06\% in 2024 showing significant improvement compared to other methods
    
    \item We conduct a comprehensive spatiotemporal analysis of oil palm dynamics, revealing a peak in plantation area in 2022 followed by a contraction in 2024. Furthermore, we quantified land cover transitions, identifying that while agricultural crop rotation remains the dominant interaction, there is a distinct and increasing trend of oil palm encroachment into flooded vegetation (wetland) areas, particularly in Kalimantan.
\end{enumerate}

The remainder of this paper is organized as follows.  We introduce the related work about semantic segmentation from remote sensing images and oil palm mapping in Sec. \ref{sec:related}.
Following that, we introduce our proposed method in detail in Sec. \ref{sec:method}. We
analyze and compare the time-series oil palm mapping results of our proposed method and other approaches in Sec. \ref{sec:exp}, followed
by comprehensive discussion and analysis in Sec. \ref{sec:diss}. Finally, we summarize this paper in Sec. \ref{sec:concl}.

\section{Related works}
\label{sec:related}
\subsection{Oil palm mapping}

The previous work related to oil palm mapping can be divided into three parts: traditional image processing methods, traditional machine learning approaches, and deep learning methods. Traditional image processing methods primarily rely on hand-crafted features and geometric rules, for example template matching \citep{templatematching}, which correlates a crown template to find trees, and morphological analysis combined with watershed segmentation \citep{shafri2011} to delineate tree boundaries. Traditional machine learning approaches have been extensively employed for land-cover classification using multi-spectral low-resolution imagery. These methods encompass random forest \citep{randomforestanddecisiontree}, XGBoost \citep{ml_ieee}, and decision tree algorithms \citep{decisiontree}. However, these approaches are highly sensitive to data quality, as label noise can significantly degrade classification performance and reduce model efficiency \citep{labelnoise}. 
With the advancement of data-driven semantic segmentation techniques, deep learning methodologies have gained considerable prominence \citep{dl_al}. These approaches have demonstrated superior performance across various remote sensing applications, including cloud shadow identification \citep{cloud}, gas quantification \citep{gasquantification}, and plant detection \citep{plantdetection}. Nevertheless, deep learning models typically require extensive training datasets to achieve optimal performance. 


\subsection{Noisy learning from low-resolution historical label}

To address the limited availability of high-resolution (HR) annotations, an increasing number of studies have explored the use of abundant low-resolution (LR) land cover products as noisy supervisory signals. Early work focused on alleviating the resolution gap by integrating multi-scale representations to smooth predictions \citep{noisy_multi}, or by casting the problem as label super-resolution through joint feature–label distribution alignment \citep{nosy_joint}.
However, directly training HR models with LR labels inevitably introduces substantial noise, motivating the development of noise-robust learning strategies. To this end, \citep{l2hnet} proposed a confident area selection mechanism to identify reliable regions within LR maps, while \citep{WESUP-LCP} introduced a weakly supervised framework that transfers high-confidence samples from LR products to generate pseudo-labels for HR training. These methods explicitly reduce the influence of mislabeled pixels and enable effective learning under imperfect supervision.
More recently, such noise-aware frameworks have been scaled to large-area and meter-level land cover mapping. \citep{SonoLC} employed self-supervised consistency losses to refine noisy constraints for 1-m resolution mapping, and \citep{SonoLCV2} further extended this paradigm through parallel feature extraction without requiring temporally aligned ground truth. 
Despite these advances, existing approaches predominantly target general land cover classes and do not explicitly account for the temporal dynamics and plantation-specific spatial patterns of oil palm systems, where the use of historical LR labels can introduce even more severe noise.

\begin{figure}[t]
    \centering
    \includegraphics[width=\linewidth]{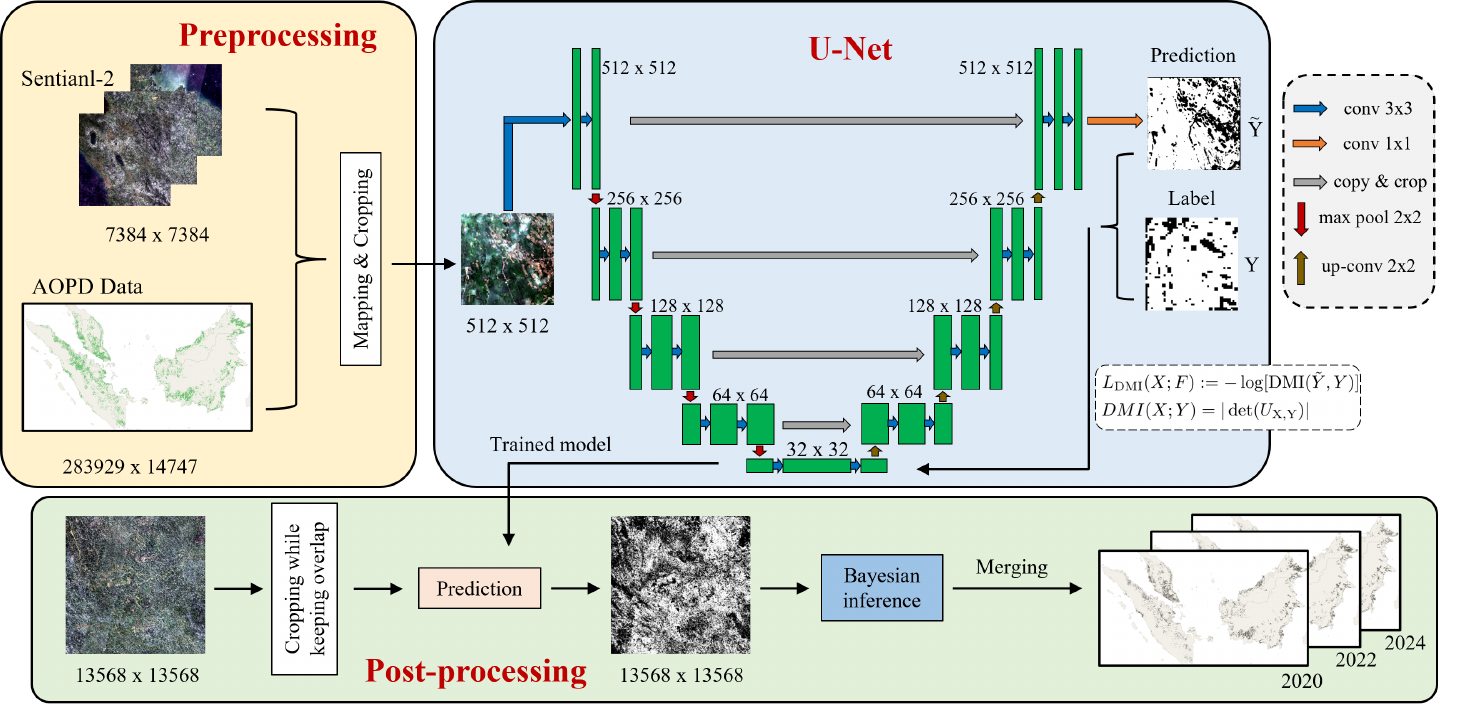}
    \caption{Process for Developing a High-Resolution Oil Palm Plantation Map}
    \label{fig:method}
\end{figure}

\section{Methodology}

\label{sec:method}

\subsection{Overview of our proposed method}

This study presents a comprehensive deep learning-based approach for mapping oil palm plantations across Indonesia and Malaysia using satellite imagery. Fig. \ref{fig:method} shows the overall flowchart of our proposed method, inlcuding four major steps, i.e., data preprocessing, neural network architecture, Determinant-based Mutual Information (DMI) loss, and post-processing for final results. We summarized these four major steps as follows:

\begin{enumerate}
    \item \textbf{Data preprocessing}. We used the Annual Oil Palm Area Dataset (AOPD) \citep{ml_annu} at 100 m resolution as noisy supervision, together with 10 m Sentinel-2 surface reflectance imagery obtained from Google Earth Engine. The imagery was cropped into 512×512 patches and split into training and validation sets.
    \item \textbf{Neural network architecture}. We evaluated five state-of-the-art semantic segmentation models—U-Net \citep{U-Net}, MMSFormer \citep{MMSFormer}, DeepLabv3+ \citep{DeepLabv3Plus}, UNetFormer \citep{UNetFormer}, and PyramidMamba \citep{PyramidMamba}—under a unified training protocol with consistent hyperparameters, including z-score normalization and SGD optimization.
    \item \textbf{DMI loss for noisy supervision}. To address the resolution mismatch between AOPD labels and Sentinel-2 imagery, we adopted the determinant-based mutual information (DMI) loss \citep{dmi_loss}, which provides robustness to noisy labels by maximizing mutual information between predictions and annotations without requiring explicit noise modeling.
    \item \textbf{Post-processing and mapping}. The trained models were applied to Sentinel-2 imagery from 2020 to 2024. Bayesian inference was used to fuse model outputs with AOPD priors, and patch-level predictions were mosaicked to produce wall-to-wall oil palm plantation maps for Indonesia and Malaysia.
\end{enumerate}

\subsection{Data preprocessing}

We obtained oil palm plantation labels at 100 meter resolution for Indonesia and Malaysia from the Annual Oil Palm Area Dataset (AOPD) \citep{ml_annu}, using data from 2016, which represents the most recent year available in this dataset. The AOPD was created using a combination of geometric techniques and machine learning approaches. For the periods 2007-2010 and 2015-2016, AOPD trained a random forest classifier with four classes (oil palm, other vegetation, water, and other) using PALSAR and PALSAR-2 parameters. 
For input imagery, we utilized Sentinel-2 surface reflectance data from the Copernicus program at 10 meter resolution, downloaded from Google Earth Engine (GEE). The imagery covers areas in Indonesia and Malaysia for 2020, which serves as the starting year for our dataset updates. We used 10 spectral bands include Blue, Green, Red, Red Edge 1-4, Near Infrared, and Short Wave Infrared 1-2.

To create a consistent dataset from the 100 meter resolution oil palm labels and 10 meter resolution satellite imagery, we resampled the oil palm labels from 100 meter to 10 meter resolution while preserving the original label values. Next, we cropped both the labels and imagery into corresponding 512×512 pixel patches based on their geographic coordinates.

To address significant class imbalance—identifiable in the initial dataset of 161,668 non-oil palm images versus only 29,568 oil palm images—we employed a random undersampling strategy. We retained all positive samples (oil palm) and randomly selected an equivalent number of negative samples (non-oil palm) to construct a balanced dataset of 59,136 images. Given that the input labels are coarse (100 meter) while the target predictions are fine-grained (10 meter), we frame this as a weakly supervised learning task; this resolution discrepancy serves to mitigate spatial autocorrelation and label leakage. The final balanced dataset was partitioned into training (70\%) and validation (30\%) subsets. Prior to model ingestion, z-score normalization was applied to all imagery to standardize pixel intensity distributions and facilitate stable convergence.

Prior to model training, we applied z-score normalization to all images to prevent training divergence. While we experimented with several deep learning architectures, U-Net provided the best accuracy, as discussed in Sec. \ref{sec:other-models}.

\subsection{Recall of U-Net}
\label{sec:unet}
The U-Net architecture was originally proposed in 2015 \citep{U-Net} for biomedical image segmentation. The original architecture consists of 18 convolutional layers with 3×3 kernels, each followed by ReLU activation functions. The network employs a symmetric encoder-decoder structure with four 2×2 max pooling operations for downsampling in the encoder path and four 2×2 upsampling layers in the decoder path. The final output layer uses a 1×1 convolution for classification. A key innovation of U-Net is the use of skip connections that concatenate feature maps from the encoder to the corresponding decoder layers, enabling the network to preserve fine-grained spatial information crucial for accurate localization.

Although originally designed for biomedical image segmentation, U-Net has proven highly effective for various segmentation tasks, including remote sensing applications as utilized in our study. Our implementation follows the original architecture with a modification, batch normalization \citep{BatchNorm} is inserted between each convolutional layer and ReLU activation function to improve training stability.
For training, we employed Stochastic Gradient Descent (SGD) as the optimizer with a learning rate of 0.001, which was the optimal hyperparameter configuration for this experiment.

\subsection{DMI Loss}
\label{sec:dmi}
Due to the resolution mismatch between oil palm plantation labels (100 meters) and both satellite images and target outputs (10 meters), we treat the training labels as noisy. To address this challenge, we employ the DMI (Determinant-based Mutual Information) loss function \citep{dmi_loss} for learning with noisy labels without requiring additional information, as shown in equation 1.

\begin{equation}
	L_{\text{DMI}}(X; F) := -\log[\text{DMI}(\tilde{Y}, Y)],
\end{equation}

where $X$ represents the input image, $F$ denotes the U-Net model, $\tilde{Y}$ is the model's prediction, and $Y$ is the ground truth. 
DMI (Determinant-based Mutual Information) \citep{dmi_ori}, which employs a DMI-Mechanism to evaluate the agreement between model predictions and labels. This mechanism encodes the relationship between predictions and labels into a 2×2 matrix $U_{\text{X,Y}}$, as defined in equation \eqref{eq:2} and \eqref{eq:3}. Each entry in $U_{\text{X,Y}}$ represents the joint occurrence probabilities of specific predicted and actual values. The determinant of this matrix quantifies the dependency structure: a positive determinant indicates positive correlation, while a negative determinant reflects negative correlation.

\begin{equation}
    DMI(X; Y) = | \det(U_{\text{X,Y}}) |,
    \label{eq:2}
\end{equation}

\begin{equation}
    U_{\text{X,Y}} = Pr[X = x, Y = y],
    \label{eq:3}
\end{equation}

where $Pr[X = x, Y = y]$ represents the matrix form of the joint probability distribution of $X$ and $Y$.

\subsection{Post-processing for final results}
We predicted oil palm plantation maps for Indonesia and Malaysia from 2020 to 2024 at 10 meter resolution using our trained model. Raw satellite imagery was downloaded from GEE for the years 2020, 2022, and 2024. To accommodate the model's input requirements, we partitioned these large images into smaller 512×512 pixel tiles with a 20 pixels overlap to ensure seamless predictions at tile boundaries.

Since the AOPD dataset \citep{ml_annu} was created using remote sensing methods with different preprocessing approaches—specifically, their model was trained on multiclass classification while ours uses binary classification, so we leveraged their product to enhance our map accuracy through Bayesian inference. This approach combines the strengths of both methods, as shown in equation 4 and 5, where the UNet prediction serves as the prior, AOPD provides the evidence, and the posterior yields a more accurate oil palm plantation map.

\begin{equation}
    p(\theta \mid \{y_{\text{i}}\}) = \frac{p(\{y_{\text{i}}\} \mid \theta)p(\theta)}{p(\{y_{\text{i}}\})},
\end{equation}

\begin{equation}
    p(\{y_i\} \mid \theta) = \prod_{i} p(y_i \mid \theta).
\end{equation}

where $\theta$ represents the probability of $y_i = 1$ (indicating oil palm presence). In this formulation, $p(\theta|{y_i})$ represents the posterior probability, $p({y_i})$ represents the evidence, $p({y_i}|\theta)$ signifies the likelihood, and $p(\theta)$ denotes the prior probability.
Finally, all 13,568×13,568 pixel images were merged to create comprehensive oil palm plantation maps for Indonesia and Malaysia, resulting in a final map size of 278,789×148,117 pixels.

\section{Experiments}
\label{sec:exp}
\subsection{Experimental setup}
Our experiments were conducted using the PyTorch open-source deep learning library on a single NVIDIA GeForce RTX 4090 GPU. We implemented a U-Net architecture \citep{U-Net} designed in Sec. \ref{sec:unet} to process input images of 512×512 pixels with 10 channeles. The model was trained from scratch using Stochastic Gradient Descent (SGD) optimizer \citep{sgd}with a momentum of 0.9. We set the learning rate to 0.001 and trained the model for 5 epochs. We utilized the DMI loss \citep{dmi_loss} as our loss function (detailed in Sec. \ref{sec:dmi}). All input images underwent z-score normalization \citep{z-score} (Eq. \eqref{eq:sigma}) to standardize intensity distributions and enhance model convergence. 

\begin{equation}
    \tilde{s}_{ki} = \frac{s_{ki} - \mu_i}{\sigma_i}
    \label{eq:sigma}
\end{equation}

where $s_{ki}$ represents the intensity value of pixel $k$ in image $i$, $\tilde{s}_{ki}$ is the corresponding normalized value, and $\mu_i$ and $\sigma_i$ are the mean and standard deviation of all pixel intensities in image $i$, respectively.

The evaluation metrics employed in this study include F1 score, User's Accuracy (UA), Producer's Accuracy (PA), and Overall Accuracy (OA). To facilitate the calculation of these metrics, we denote the number of correctly identified positive pixels as True Positives ($TP$), incorrectly identified positive pixels as False Positives ($FP$), correctly identified negative pixels as True Negatives ($TN$), and incorrectly identified negative pixels as False Negatives ($FN$).

Precision (equivalent to User's Accuracy) indicates the proportion of positive predictions that were actually correct, while Recall (equivalent to Producer's Accuracy) measures the proportion of actual positive cases that were correctly identified. These metrics are defined in Eq. \eqref{eq:7} and \eqref{eq:8}, respectively.

\begin{equation}
    \text{Precision} = \frac{TP}{TP + FP}
    \label{eq:7}
\end{equation}

\begin{equation}
    \text{Recall} = \frac{TP}{TP + FN}
    \label{eq:8}
\end{equation}

The F1 score evaluates the classification model's performance by calculating the harmonic mean of precision and recall, as shown in Eq. \eqref{eq:9}. This metric assesses the balance between identifying positive cases and minimizing false detections.

\begin{equation}
    \text{F1 Score} = 2 \times \frac{\text{Precision} \times \text{Recall}}{\text{Precision} + \text{Recall}}
    \label{eq:9}
\end{equation}

User's Accuracy and Producer's Accuracy correspond conceptually to precision and recall but are expressed as percentages, as demonstrated in Eq. \eqref{eq:10} and \eqref{eq:11}. Overall Accuracy (Eq. \eqref{eq:12}) represents the percentage of all predictions that were correct across both classes.

\begin{equation}
    \text{User's Accuracy} = \frac{TP}{TP + FP} \times 100\%
    \label{eq:10}
\end{equation}

\begin{equation}
    \text{Producer's Accuracy} = \frac{TP}{TP + FN} \times 100\%
    \label{eq:11}
\end{equation}

\begin{equation}
    \text{Overall Accuracy} = \frac{TP + TN}{TP + TN + FP + FN} \times 100\%
    \label{eq:12}
\end{equation}

\subsection{Generation of multi-temporal validation data}
\begin{figure}[t]
    \centering
    \includegraphics[width=\linewidth]{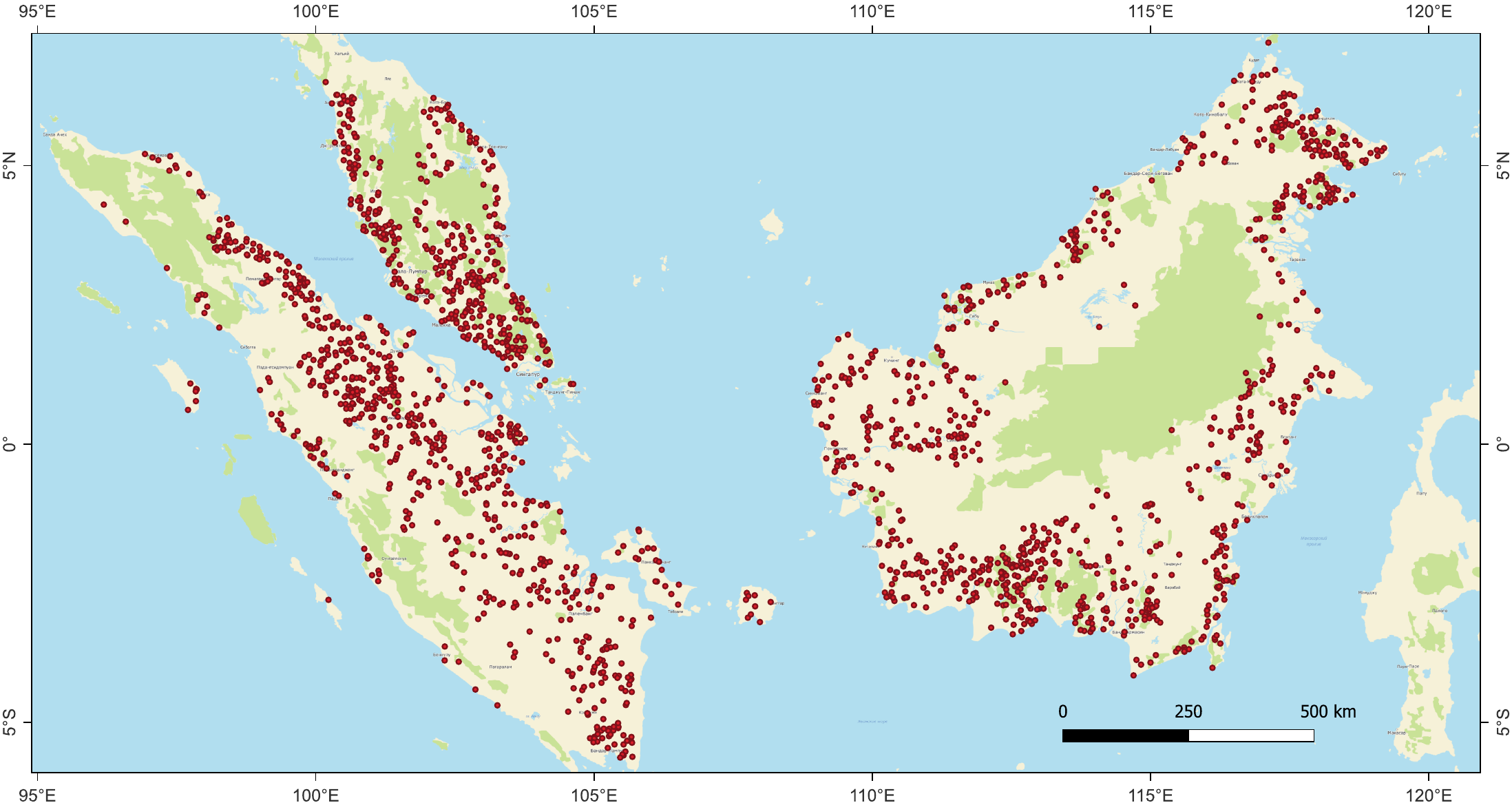}
    \caption{Geographic distribution of the 2,058 validation points used for accuracy assessment across the study area.}
    \label{fig:validation_points}
\end{figure}

To rigorously evaluate the performance of the proposed models and the AOPD data, we established a validation set comprising 2,058 randomly sampled pixels spanning Indonesia and Malaysia (Fig. \ref{fig:validation_points}). To ensure temporal comparability in the assessment, the identical set of spatial coordinates was utilized for the validation of the 2020, 2022, and 2024 images.

\subsection{Other comparative deep learning-based methods}
\label{sec:other-models}
In addition to the U-Net model, this study evaluates several state-of-the-art deep learning methods for semantic segmentation, including MMSFormer~\citep{MMSFormer}, DeepLabv3+~\citep{DeepLabv3Plus}, UNetFormer~\citep{UNetFormer}, HRNet~\citep{HRNet}, and PyramidMamba~\citep{PyramidMamba}. All these methods are recognized for their efficiency in semantic segmentation tasks.

\begin{enumerate}
    \item MMSFormer \citep{MMSFormer}: A transformer-based architecture designed for multi-modal fusion, utilizing Mix-Transformer encoders and parallel multi-scale convolutions.
    \item DeepLabv3+ \citep{DeepLabv3Plus}: An encoder-decoder network that uses Atrous Spatial Pyramid Pooling (ASPP) to refine segmentation boundaries and capture contextual information.
    \item UNetFormer \citep{UNetFormer}: A hybrid model integrating a ResNet encoder with a transformer decoder to capture both long-range dependencies (via self-attention) and local details.
    \item HRNet \citep{HRNet}: A high-resolution network that connects parallel multi-resolution streams to preserve spatial details without recovering resolution from downsampled features.
    \item PyramidMamba \citep{PyramidMamba}: A novel architecture incorporating Mamba-based decoders and Dense Spatial Pyramid Pooling to efficiently process spatial contexts with reduced redundancy.
\end{enumerate}

\begin{figure}[t]
    \centering
    \includegraphics[width=\linewidth]{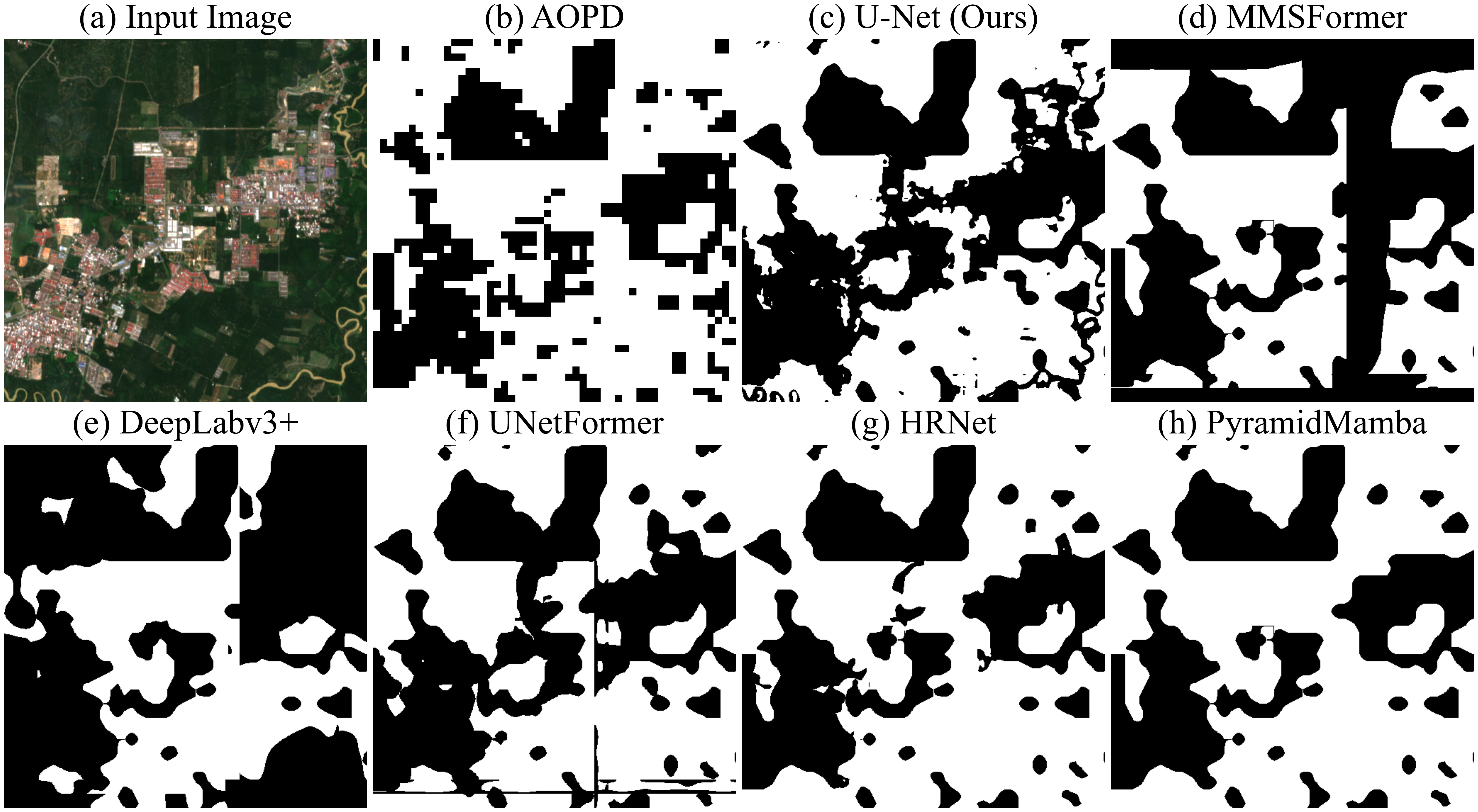}
    \caption{Oil palm detection results in Malaysia across different models: (a) Sentinel-2 imagery, (b) AOPD noisy label data used for model training, and (c)-(h) oil palm detection outputs from respective models}
    \label{fig:MaMap}
\end{figure}

\begin{figure}[t]
    \centering
    \includegraphics[width=\linewidth]{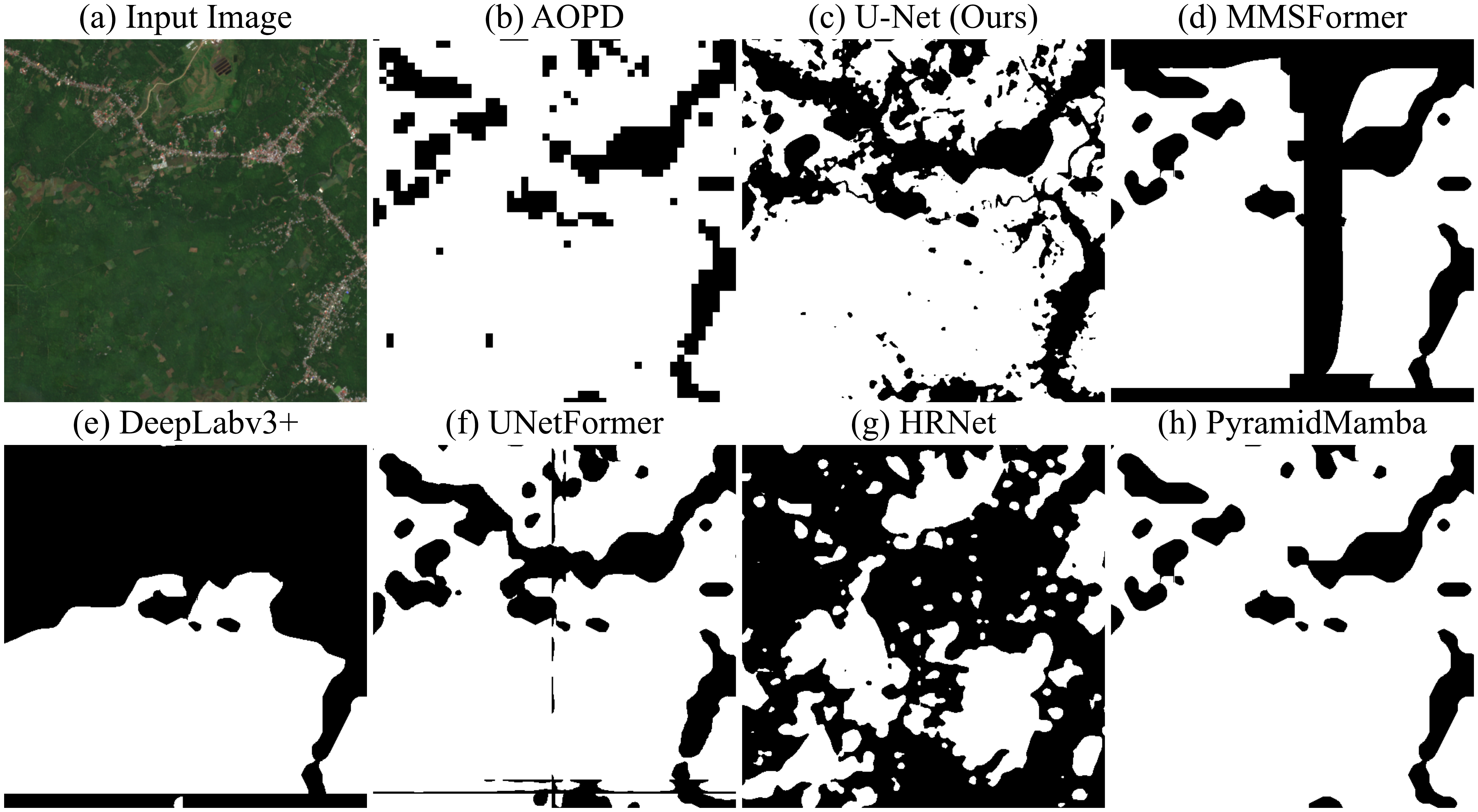}
    \caption{Oil palm detection results in Indonesia across different models: (a) Sentinel-2 imagery, (b) AOPD noisy label data used for model training, and (c)-(h) oil palm detection outputs from respective models}
    \label{fig:InMap}
\end{figure}

\subsection{Experimental results}

Visual inspection (Fig.  \ref{fig:MaMap} and \ref{fig:InMap}) reveals that the U-Net model yields the most precise boundary delineation among all evaluated methods. The model successfully recovers fine-grained morphological details, such as the separation between plantation blocks and adjacent roads or rivers, which are notably absent in the coarse 100-meter AOPD training labels. In contrast, while newer architectures like UNetFormer and PyramidMamba produce smoother outputs, they tend to over-smooth boundaries, effectively replicating the blocky artifacts of the noisy labels rather than learning the true underlying features. Furthermore, models such as MMSFormer and DeepLabv3+ exhibited performance inferior to the baseline AOPD data, suggesting a limited capacity to generalize from noisy supervision in this specific domain.

\begin{table}[t]
\label{tab:performance_comparison}
\centering
\caption{Accuracy comparison between AOPD, U-Net, MMSFormer, DeepLabv3+, UNetFormer, HRNet and PyramidMamba in 2020, 2022 and 2024 year}
\scriptsize
\begin{tabular}{l|cccc|cccc|cccc}
\toprule
\multirow{2}{*}{Method} & \multicolumn{4}{c|}{2020} & \multicolumn{4}{c|}{2022} & \multicolumn{4}{c}{2024} \\
\cmidrule(r){2-5} \cmidrule(r){6-9} \cmidrule(r){10-13}
 & F1 & UA & PA & OA & F1 & UA & PA & OA & F1 & UA & PA & OA \\
\midrule
AOPD          & 0.49 & 34.10  & 86.94 & 35.93 & 0.54 & 39.22 & 88.72 & 40.67 & 0.55 & 40.23 & 88.84 & 41.52 \\
U-Net         & \textcolor{red}{0.69} & \textcolor{red}{55.03} & 93.13 & \textcolor{red}{70.64} & 0.62 & \textcolor{red}{53.04} & 74.54 & \textcolor{red}{63.53} & 0.58 & \textcolor{red}{50.84} & 67.26 & \textcolor{red}{60.06} \\
MMSFormer     & 0.50 & 37.54 & 74.05 & 47.23 & 0.54 & 42.76 & 74.70 & 50.03 & 0.55 & 43.41 & 74.55 & 49.91 \\
DeepLabv3+    & 0.48 & 40.19 & 59.45 & 54.35 & 0.50 & 44.30 & 56.25 & 54.35 & 0.53 & 47.05 & 59.38 & 56.11 \\
UNetFormer    & 0.63 & 48.20 & 91.92 & 62.19 & \textcolor{red}{0.64} & 50.22 & 87.20 & 60.43 & \textcolor{red}{0.62} & 50.14 & 80.65 & 59.33 \\
HRNet         & 0.62 & 46.35 & 95.02 & 59.33 & 0.63 & 49.02 & 87.65 & 58.72 & 0.61 & 48.57 & 83.63 & 57.14 \\
PyramidMamba  & 0.54 & 37.87 & \textcolor{red}{95.53} & 42.98 & 0.59 & 42.64 & \textcolor{red}{95.43} & 46.99 & 0.60 & 43.60 & \textcolor{red}{95.24} & 47.72 \\
\bottomrule
\end{tabular}
\label{tab:performance_comparison}
\end{table}

To ensure a fair and unbiased assessment—given the noisy nature of the training labels—we utilized an independent validation strategy. We conducted a manual visual interpretation of the 2,058 sampled pixels (Fig. \ref{fig:validation_points}) using Google Earth imagery to establish ground truth for 2020, 2022, and 2024. Based on these verified labels, we computed F1 Score, UA, PA, and OA. This method enables a robust comparison of the U-Net and AOPD baselines against the five advanced models (Table \ref{tab:performance_comparison}), effectively bypassing the inaccuracies present in the training ground truth.

The results demonstrate that U-Net consistently achieves the highest performance across most evaluation metrics throughout the study period. Notably, UNetFormer exhibits superior F1 scores in 2022 and 2024, indicating its effectiveness in balancing precision and recall. Furthermore, PyramidMamba consistently maintains the highest Producer's Accuracy values across all three years, demonstrating its exceptional capability in correctly identifying positive instances within the dataset. 

\subsection{Ablation Studies}
\begin{figure}[t]
    \centering
    \includegraphics[width=\linewidth]{imgs/ablation_studies.pdf}
    \caption{Visual comparison between (a) Sentinel-2 satellite imagery, (b) Ground Truth (AOPD), (c) Results from the U-Net model trained with DMI Loss (learning with noisy labels), and (d) Results from the U-Net model trained with Binary Cross-Entropy Loss (standard supervision).}
    \label{fig:ablation}
\end{figure}

To illustrate the significant challenge posed by the resolution mismatch between high-resolution imagery (10 meters) and spatially coarse labels (100 meters), and to demonstrate the efficacy of our learning with noisy labels approach, we conducted an ablation study. Fig.  \ref{fig:ablation} presents a qualitative comparison of a UNet model trained under identical configurations, differing only in the loss function employed. 

Panel (c) displays the results using DMI Loss \citep{dmi_loss}, the technique adopted in this study for learning with noisy labels. In contrast, panel (d) shows the results using Binary Cross-Entropy (BCE) loss \citep{bce}, representing the standard supervised training baseline. The difference in performance is substantial; the UNet trained with standard BCE loss completely fails to detect oil palm features within the Sentinel-2 imagery. Conversely, the model trained with DMI loss not only detects oil palm with high precision but also recovers granular spatial details that surpass the coarse resolution of the original ground truth (AOPD \citep{ml_annu}).

To further substantiate these qualitative observations with quantitative evidence, Table \ref{tab:ablation_result} provides a numerical performance comparison. These metrics clearly demonstrate that the model trained with DMI Loss consistently outperforms the BCE Loss baseline across all evaluated years.

\section{Discussions}
\label{sec:diss}

\begin{table}[t]
    \centering
    \caption{Accuracy comparison between U-Net trained with DMI loss and BCE loss across different years.}
    \label{tab:ablation_result}
    \footnotesize
    \begin{tabular}{cc|cc}
        \toprule
        \multirow{2}{*}{Year} & \multirow{2}{*}{Metric} & \multicolumn{2}{c}{Method} \\
        \cmidrule(l){3-4}
         & & U-Net (DMI Loss) & U-Net (BCE Loss) \\
        \midrule
        \multirow{4}{*}{2020} 
         & F1 & \textcolor{red}{0.69} & 0.50 \\
         & UA & \textcolor{red}{55.03} & 37.54 \\
         & PA & \textcolor{red}{93.13} & 74.05 \\
         & OA & \textcolor{red}{70.64} & 47.23 \\
        \midrule
        \multirow{4}{*}{2022} 
         & F1 & \textcolor{red}{0.62} & 0.54 \\
         & UA & \textcolor{red}{53.04} & 42.76 \\
         & PA & 74.54 & \textcolor{red}{74.70} \\
         & OA & \textcolor{red}{63.53} & 50.03 \\
        \midrule
        \multirow{4}{*}{2024} 
         & F1 & \textcolor{red}{0.58} & 0.55 \\
         & UA & \textcolor{red}{50.84} & 43.41 \\
         & PA & 67.26 & \textcolor{red}{74.55} \\
         & OA & \textcolor{red}{60.06} & 49.91 \\
        \bottomrule
    \end{tabular}
\end{table}

\subsection{Comparison of our results with statistics and other products}

We first check the differences between Predicted Data and Statistical Data by Spearman's rank correlation coefficient,The general formula that works for all cases, including when there are tied ranks:

\begin{equation}
\rho = \frac{\frac{1}{n} \sum_{i=1}^{n} \left( R(x_i) - \overline{R(x)} \right) \cdot \left( R(y_i) - \overline{R(y)} \right)}{\sqrt{\left( \frac{1}{n} \sum_{i=1}^{n} \left( R(x_i) - \overline{R(x)} \right)^2 \right) \cdot \left( \frac{1}{n} \sum_{i=1}^{n} \left( R(y_i) - \overline{R(y)} \right)^2 \right)}}
\label{eq:correlation_coefficient}
\end{equation}

where $x_i$ and $y_i$ are the $i$-th paired observations, $n$ denotes the total number of observations, and $\overline{R(x)}$ and $\overline{R(y)}$ are the average ranks of variables $x$ and $y$, respectively. For data without tied ranks, both mean ranks equal $\frac{n+1}{2}$.

\begin{figure}[b]
    \centering
    \includegraphics[width=\linewidth]{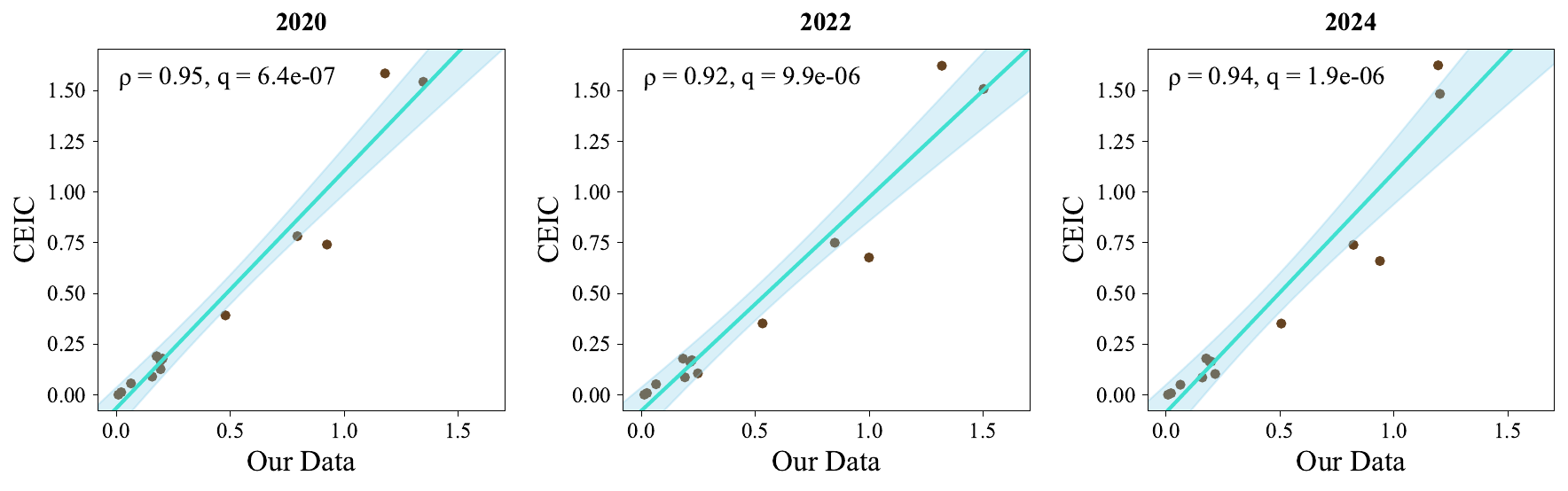}
    \caption{Temporal Validation of Predictive Model Performance (2020-2024) with CEIC dataset}
    \label{fig:statistic}
\end{figure}

We validated the temporal consistency of our mapping results by comparing the predicted plantation areas against the Global Economic Data (CEIC) \citep{ceic2023} using Spearman's rank correlation coefficient (Eq.  \eqref{eq:correlation_coefficient}). As illustrated in Fig.  \ref{fig:statistic}, our predictions demonstrate a strong positive correlation with the reference statistical data across all three years, yielding coefficients of 0.95, 0.92, and 0.94 for 2020, 2022, and 2024, respectively. Furthermore, the calculated q-values (6.4e-07, 9.9e-06, and 1.9e-06) fall well below the standard significance threshold ($\alpha = 0.05$). These results provide robust statistical evidence that our model maintains excellent rank-order agreement with official economic statistics, confirming the reliability of the derived trends despite the absence of dense manual annotations.

\begin{figure}[t]
    \centering
    \includegraphics[width=\linewidth]{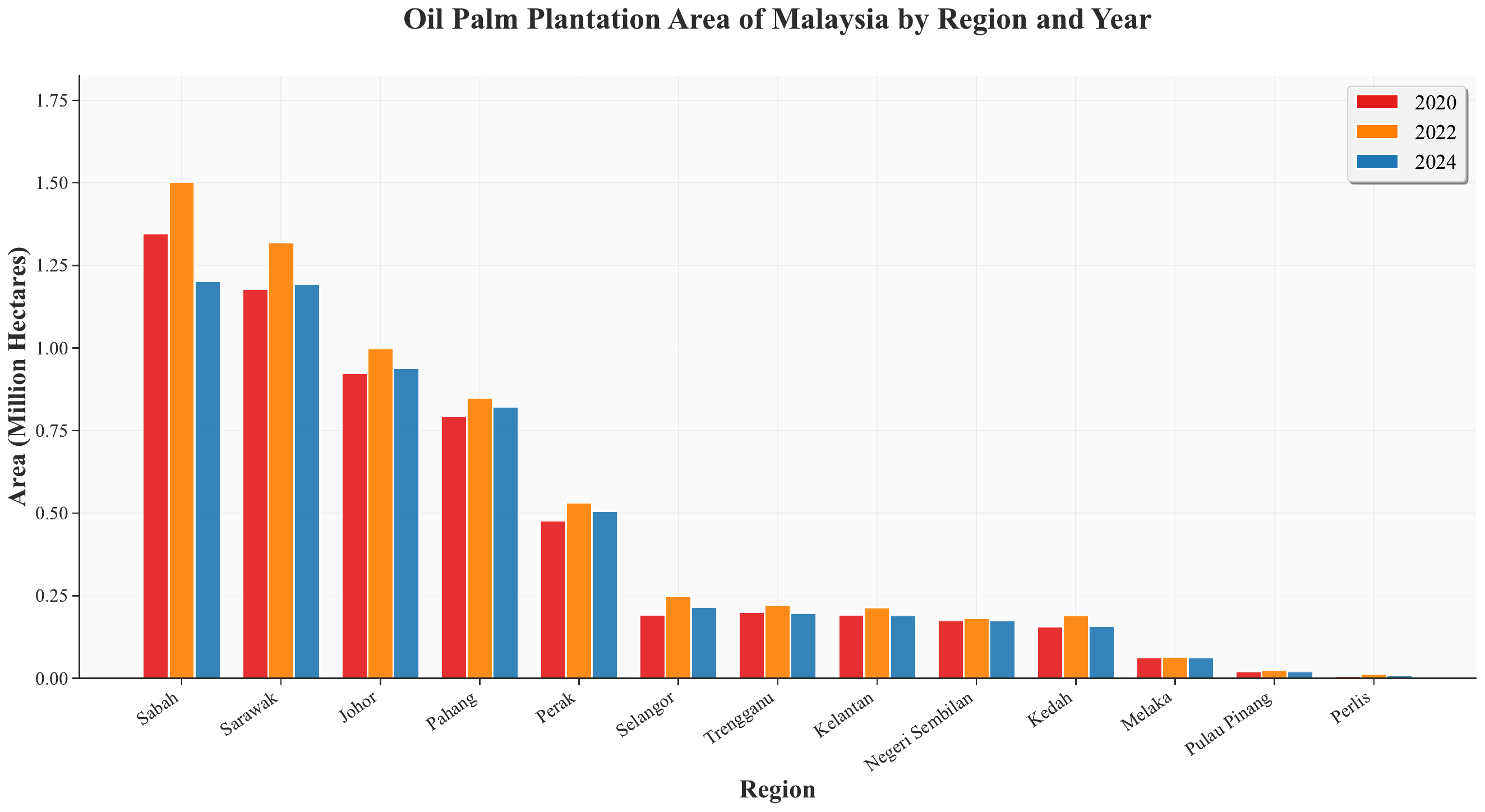}
    \caption{Oil Palm Plantation Area of Malaysia by Region and Year}
    \label{fig:oil_Malay}
\end{figure}

\begin{figure}[t]
    \centering
    \includegraphics[width=\linewidth]{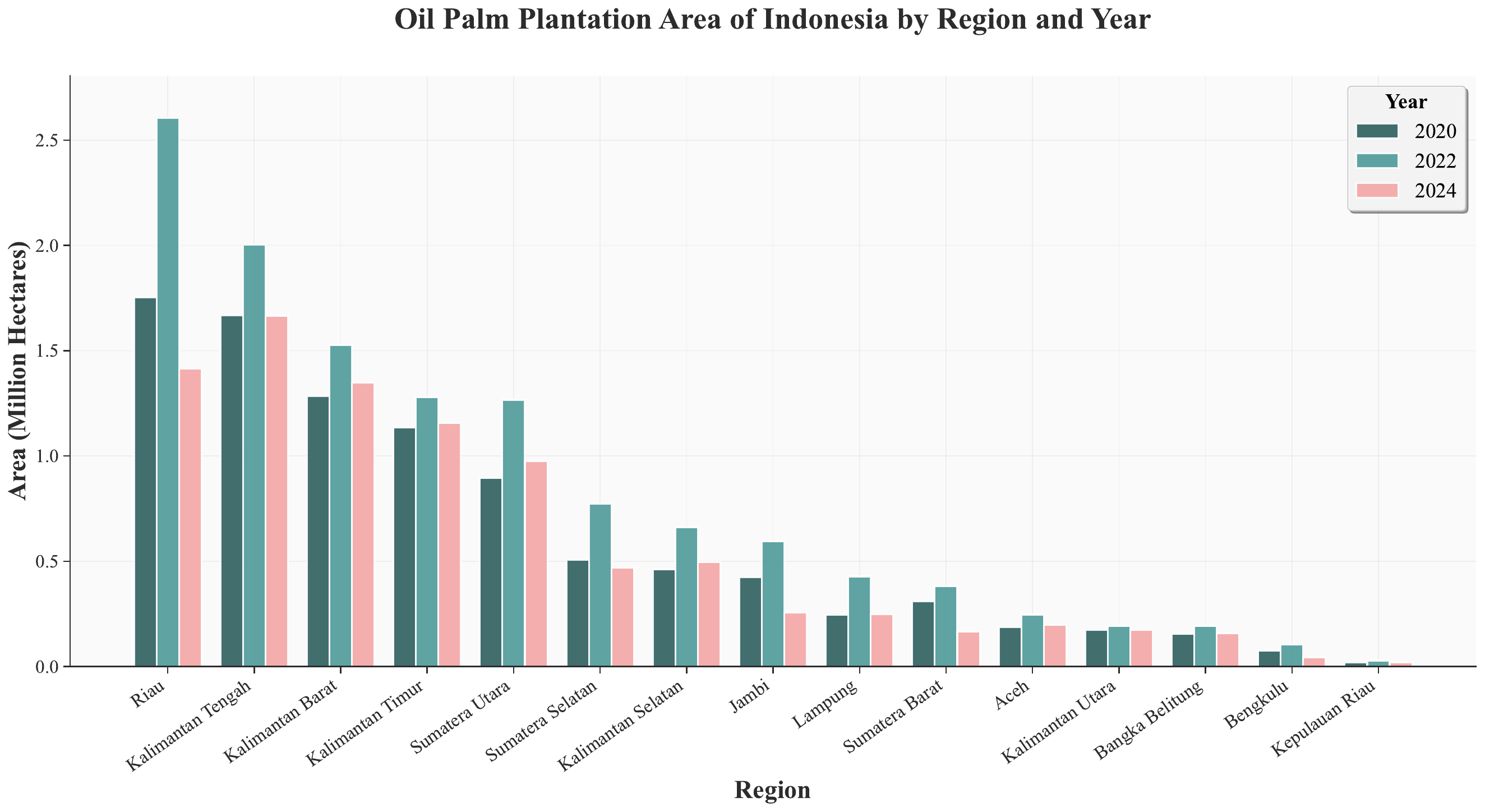}
    \caption{Oil Palm Plantation Area of Indonesia by Region and Year}
    \label{fig:oil_Indonsia}
\end{figure}


Fig.  \ref{fig:oil_Malay} presents the oil palm plantation area across 13 Malaysian regions for 2020, 2022, and 2024. Overall, the areas remained relatively stable during the study period, consistent with the maturity of Malaysia’s oil palm sector. Sabah and Sarawak consistently ranked as the top-producing states, with cultivated areas ranging from $1.1 \times 10^{6}$ ha to $1.5 \times 10^{6}$ ha, reflecting their dominant role in national production \citep{plam_area}. Johor, Pahang, and Perak followed, maintaining substantial and stable extents between $0.5 \times 10^{6}$ ha and $1.0 \times 10^{6}$ ha. While most regions showed little change, Sabah and Sarawak exhibited a notable decline of approximately $1.0 \times 10^{5}$ ha to $2.5 \times 10^{5}$ ha between 2022 and 2024, potentially due to policy adjustments, land-use reallocation, or market-driven replanting cycles.

\begin{figure}[t]
    \centering
    \includegraphics[width=\linewidth]{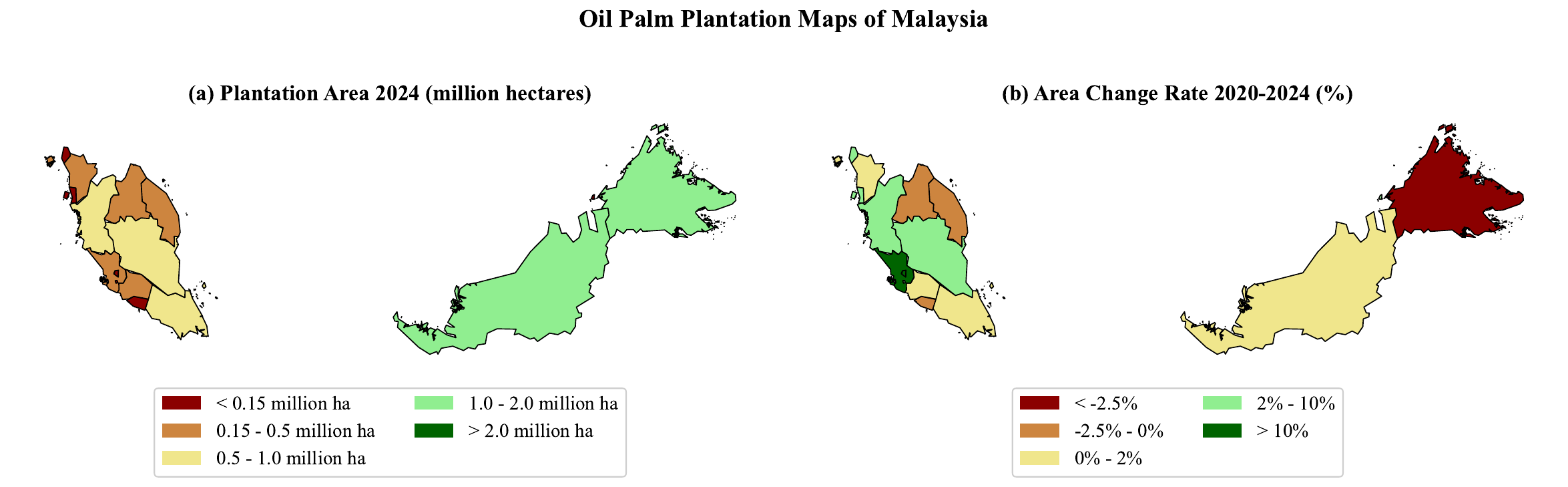}
    \caption{Oil Palm Plantation Maps in Malaysia (a) Plantation area by region in 2024 and (b) Percentage change in plantation area between 2020 and 2024}
    \label{fig:oil_maps_Malay}
\end{figure}


Similarly, Fig.  \ref{fig:oil_Indonsia} shows the spatial distribution of oil palm plantations across 15 Indonesian regions from 2020 to 2024. Riau consistently had the largest area, peaking at $2.54 \times 10^6$ ha in 2022, but declined sharply to $1.45 \times 10^6$ ha by 2024, with a reduction of ~43\%. Plantation coverage region-wide peaked in 2022, followed by a notable contraction in 2024, likely due to environmental degradation, policy restrictions on expansion, shifts in market demand, or economic factors such as commodity price fluctuations and trade regulations. Further investigation would be necessary to determine the primary drivers behind this substantial reduction in oil palm cultivation.


\subsection{Spatiotemporal Distribution and Regional Trends}

This study presents oil palm plantation maps for Malaysia and Indonesia from 2020 to 2024. Fig.  \ref{fig:oil_maps_Malay} reveals that plantations in Malaysia are predominantly concentrated in northern Kalimantan (Malaysian Borneo), with areas ranging from $1.0 \times 10^{6}$ to $2.0 \times 10^{6}$ ha, reflecting favorable agro-ecological conditions and extensive land availability \citep{malay_oil_palm_1,malay_oil_palm_2}. In contrast, the Malay Peninsula shows lower densities, averaging ~$0.65 \times 10^{6}$ ha, due to more limited expansion space and less suitable soils. The spatial disparity highlights the influence of climate, soil, and historical land-use patterns. Recent trends also indicate faster land change rates on the Malay Peninsula (2\%–10\% annually) compared to Kalimantan (−2.5\%–2\%), possibly linked to the expansion of oil palm processing and encroachment into undeveloped areas following plantation saturation \citep{malay_oil_palm_3}.

\begin{figure}[t]
    \centering
    \includegraphics[width=\linewidth]{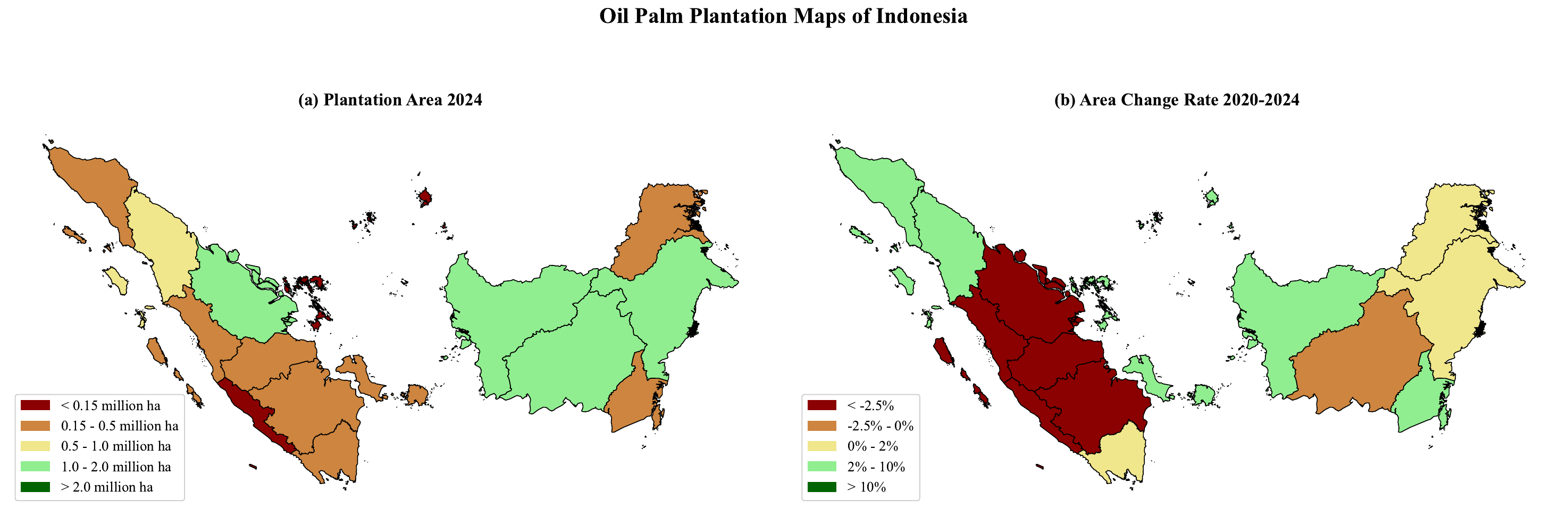}
    \caption{Oil Palm Plantation Maps in Indonesia (a) Plantation area by region in 2024 and (b) Percentage change in plantation area between 2020 and 2024}
    \label{fig:oil_maps_Indonesia}
\end{figure}

Fig.  \ref{fig:oil_maps_Indonesia} shows that oil palm plantations in Indonesia are heavily concentrated in Kalimantan, averaging $1.8 \times 10^{6}$ ha and forming the core of national production. This region is also the main driver of expansion, with an average growth rate of 5\%, while other areas declined by ~1\%. In contrast, Sumatra has extensive plantation coverage—exceeding 2.0 million hectares in some provinces—but shows stagnant or declining trends (–2.5\% to 0\%), suggesting land saturation or stricter regulations. Java and eastern islands have minimal coverage (typically <0.15 million hectares) and mixed growth, reflecting their limited role in production. The regional divergence between expansion in Kalimantan and contraction elsewhere highlights the influence of environmental, economic, and policy factors shaping Indonesia’s oil palm landscape.


\begin{figure}[t]
    \centering
    \includegraphics[width=\linewidth]{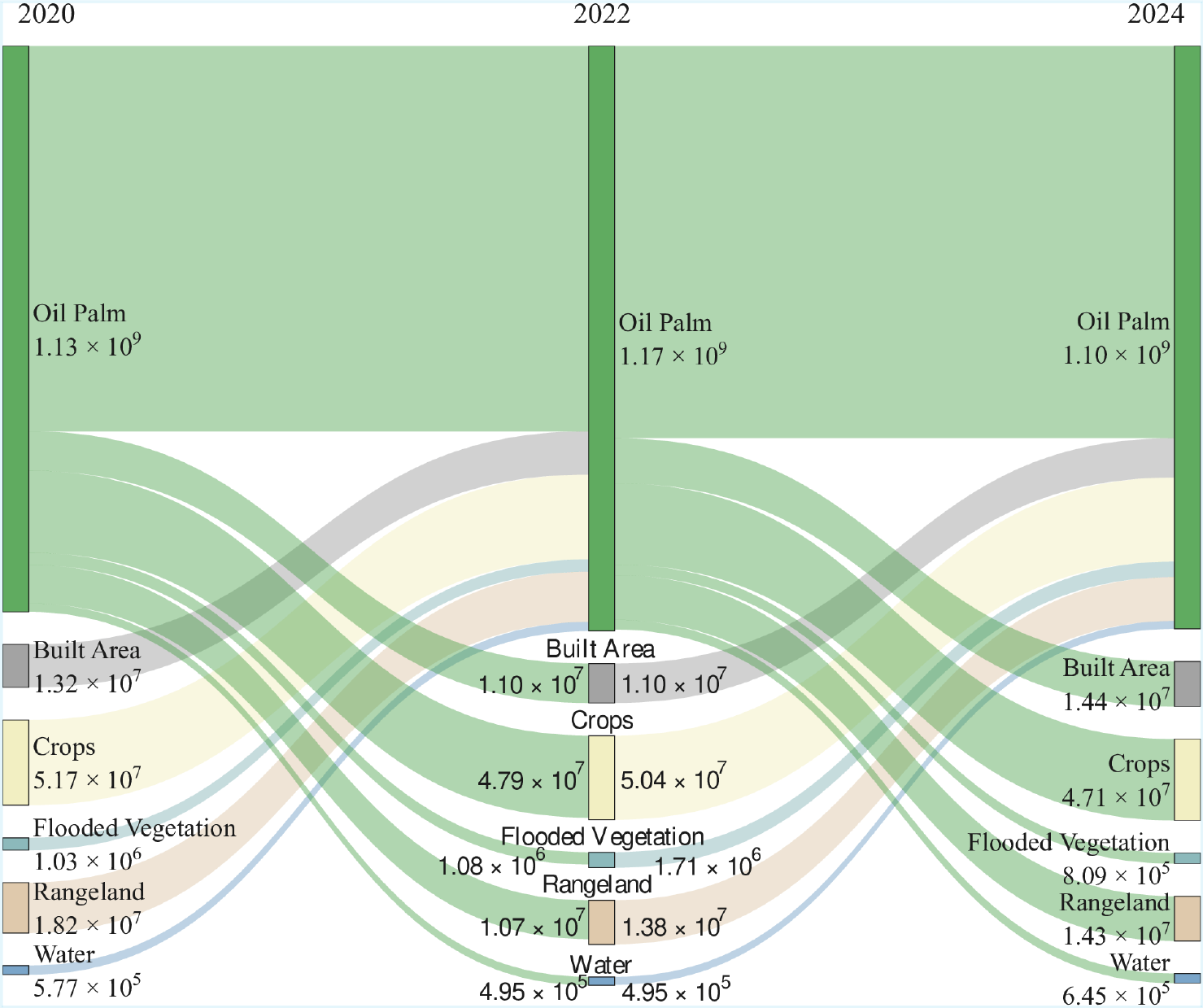}
    \caption{Sankey diagram illustrating land cover transitions (in 10-meter resolution pixels) between oil palm and other classes, including built area, crops, flooded vegetation, rangeland, and water.}
    \label{fig:oil_palm_change}
\end{figure}

\subsection{Land Cover Transition and Conversion Dynamics}
Based on the generated 10-meter resolution oil palm plantation maps, we analyzed the spatiotemporal dynamics of oil palm cultivation from 2020 to 2024. To characterize the land cover types interacting with oil palm, we utilized the Environmental Systems Research Institute (Esri) Land Cover Classification \citep{Esri_LULC_2024}. In the land cover flow analysis presented in Fig.  \ref{fig:oil_palm_change}. The total oil palm area initially expanded from $1.13 \times 10^9$ pixels in 2020 to $1.17 \times 10^9$ in 2022, driven by significant gains from other land types. However, this trend reversed in the subsequent period, with the total area contracting to $1.10 \times 10^9$ pixels in 2024, reflecting a shift in conversion momentum.

Transitions involving built area, rangeland, and water indicated a growing trend of plantations being repurposed or abandoned. Between the two intervals (2020--2022 vs. 2022--2024), the rate of land conversion into oil palm from these classes decreased; for example, built area conversion dropped from $1.32 \times 10^7$ to $1.10 \times 10^7$. Conversely, the loss of oil palm to these classes accelerated, with conversion to built area rising from $1.10 \times 10^7$ to $1.44 \times 10^7$. This trajectory suggests that oil palm expansion is slowing in these sectors in favor of urbanization and reversion to rangeland or water bodies.

In contrast, interactions with flooded vegetation and crops followed distinct patterns. Flooded vegetation exhibited an inverse trend indicative of encroachment into wetlands, where conversion into oil palm increased from $1.03 \times 10^6$ to $1.71 \times 10^6$, while reversion to flooded vegetation declined. Meanwhile, the exchange between oil palm and crops showed a reduction in bidirectional flow, suggesting a stabilization in agricultural rotation. Despite this reduction, the crop-to-palm interaction remained the most substantial, accounting for the highest proportion of land cover change among all analyzed classes.

\subsection{Uncertainty of our results}

Although our proposed method significantly improves spatial resolution from 100 meters to 10 meters, several sources of uncertainty remain in the final oil palm plantation maps. The primary source of uncertainty stems from the propagation of errors inherent in the training labels. As noted in the AOPD documentation \citep{ml_annu}, the original 100-meter datasets contain confusion between oil palm and other vegetation types, particularly in impervious areas and mixed-crop plantations. While the DMI loss function \citep{dmi_loss} effectively mitigates the impact of noisy labels during training, it cannot fully eliminate the bias introduced when the ground truth itself contains systematic classification errors.

Secondly, the exclusive reliance on optical Sentinel-2 imagery introduces susceptibility to atmospheric interference, which is a pervasive challenge in the tropical climates of Indonesia and Malaysia. Although we utilized surface reflectance products, persistent cloud cover, haze, and cloud shadows can distort the spectral reflectance values of the bands. In the training phase, this spectral noise may prevent the model from learning a consistent feature representation for oil palm. In the prediction phase, residual thin clouds or unmasked shadows can alter the radiometric integrity of pixels, potentially leading to errors in persistently cloudy regions.

Thirdly, spectral confusion remains a challenge in optical remote sensing \citep{ml_in}. Mature oil palm trees share similar spectral signatures with other tropical perennial crops, most notably coconut palms and rubber plantations . In regions like coastal Indonesia, where coconut and oil palm plantations frequently coexist, the model may generate false positives. This uncertainty is exacerbated in smallholder plantations where crops are often intercropped rather than planted in distinct, large-scale monocultures.

Finally, uncertainty is introduced through the resolution mismatch during the validation process. Our model predicts boundaries at a 10-meter granularity, capturing small roads and drainage canals within plantations. However, comparing these fine-grained predictions against 100-meter resolution historical data or statistical aggregates often results in a “mixed pixel” problem, where the coarse resolution of the reference data fails to capture the fragmented nature of plantation edges, potentially skewing the accuracy metrics in transition zones.




\section{Conclusions}
\label{sec:concl}

This study addressed the critical gap in high-resolution, up-to-date monitoring of oil palm plantations across Malaysia and Indonesia. By leveraging a noise-robust deep learning framework, we successfully generated 10-meter resolution maps for the years 2020, 2022, and 2024, overcoming the significant challenge of supervising high-resolution Sentinel-2 imagery with coarse 100-meter historical labels. Our experimental results demonstrated that the U-Net architecture, trained with DMI loss, exhibited superior robustness to label noise compared to more complex state-of-the-art models such as MMSFormer and UNetFormer, achieving a maximum overall accuracy of 70.64\% in 2020.

The spatiotemporal analysis derived from these maps provided significant insights into regional plantation dynamics. We observed a distinct fluctuation in total cultivated area, characterized by a peak expansion in 2022 followed by a notable contraction in 2024. Regionally, the expansion remains concentrated in Kalimantan, while Sumatra exhibits signs of saturation or decline. Crucially, our land cover transition analysis identified that while the conversion between oil palm and general crops constitutes the largest volume of change, the conversion of flooded vegetation (wetlands) into oil palm plantations has intensified between 2022 and 2024. This finding signals potential ecological risks despite the overall reduction in total plantation area. These generated maps serve as a vital resource for policymakers and environmental stakeholders, offering a granular view of recent land-use shifts that is essential for enforcing zero-deforestation commitments and managing sustainable agricultural development.

\section*{Data availability}

The raw data used in this study are publicly available from Sentinel-2 multispectral imagery (\url{https://dataspace.copernicus.eu/data-collections/copernicus-sentinel-data/sentinel-2}). 
The processed results generated in this study, the nationwide 10 m oil palm mapping for Malaysia and Indonesia (2022–2024), are publicly available from Zenodo under a CC-BY 4.0 licence, at DOI: \url{https://doi.org/10.5281/zenodo.17768444} \citep{kuapanich_2025_17768444}.  Any additional information or data used in this study are available from the corresponding author upon reasonable request.

\section*{Acknowledgements}

This work was supported by the National Natural Science Foundation of China (Grant No. 42401415 and No. T2125006), Fundamental Research Funds for the Central Universities, Sun Yat-sen University (Project 24xkjc002), and Jiangsu Innovation Capacity Building ProgramProject No.BM2022028).

\bibliography{main}

\end{document}